\documentclass[twocolumn,natbib]{svjour3}          
\smartqed  
\usepackage{graphicx}
\usepackage[utf8]{inputenc}
\usepackage{graphicx}
\usepackage{mathrsfs}
\usepackage{amssymb}
\usepackage{commath}
\usepackage{amsmath}
\usepackage{caption}
\usepackage{subcaption}
\usepackage{todonotes}
\usepackage{boxedminipage}
\usepackage{alltt}
\begin{document}

\title{Semantic navigation with domain knowledge}


\author{Rafael Gomes Braga$^1$         \and
        Sina Karimi$^2$                \and
        Ulrich Dah-Achinanon$^3$       \and
        Ivanka Iordanova$^2$           \and
        David St-Onge$^1$
}


\institute{
    Rafael Gomes Braga  \at
      \email{rafael.gomes-braga.1@ens.etsmtl.ca}
    \and
    Sina Karimi \at
      \email{sina.karimi.1@ens.etsmtl.ca}
    \and
    Ulrich Dah-Achinanon \at
      \email{ulrich.dah-achinanon@polymtl.ca}
    \and
    Ivanka Iordanova \at
      \email{ivanka.iordanova@etsmtl.ca}
    \and
    David St-Onge \at
      \email{david.st-onge@etsmtl.ca}
    \and
    $^1$ École de Technologie Supérieure, Mechanical Engineering dept. 1100 Notre-Dame St W, Montréal, QC H3C 1K3
    \\
    $^2$ École de Technologie Supérieure, Construction Engineering dept. 1100 Notre-Dame St W, Montréal, QC H3C 1K3
    \\
    $^3$ École Polytechnique de Montréal, 2900 Boulevard Edouard-Montpetit, Montréal, QC H3T 1J4, Canada
}

\date{Received: date / Accepted: date}

\maketitle

\begin{abstract}
Several deployment locations of mobile robotic systems are human made (i.e. urban firefighter, building inspection, property security) and the manager may have access to domain-specific knowledge about the place, which can provide semantic contextual information allowing better reasoning and decision making. In this paper we propose a system that allows a mobile robot to operate in a location-aware and operator-friendly way, by leveraging semantic information from the deployment location and integrating it to the robot’s localization and navigation systems. We integrate Building Information Models (BIM) into the Robotic Operating System (ROS), to generate topological and metric maps fed to an layered path planner (global and local). A map merging algorithm integrates newly discovered obstacles into the metric map, while a UWB-based localization system detects equipment to be registered back into the semantic database. The results are validated in simulation and real-life deployments in buildings and construction sites.
\end{abstract}

\section{Introduction}
\label{sec:introduction}
As mobile robots are deployed in the real world to perform high level tasks in complex scenarios, possibly around humans, it becomes increasingly interesting to integrate rich data from the environment into the robot's systems. While current navigation techniques based on geometric information allow the robot to navigate safely and efficiently in challenging environments, applications such as self-driving vehicles, service robots and automated construction progress monitoring require a higher level of reasoning and decision making by the robot. For example, a robot's task could be ``navigate from the hall to the kitchen", in which case a framework that integrates the room identity along with the geometric information is required.

The relatively new field of semantic mapping tackles this problem by investigating ways to generate maps that contain the geometric and semantic information of the environment, usually combining semantic segmentation techniques with SLAM algorithms. Semantic information is extracted from sensors such as depth cameras in \cite{kochanov2016scene} and LiDAR in \cite{chen2019suma++}. Most current research employs deep learning methods, for example \cite{mccormac2017semanticfusion} and \cite{xiang2017rnn}, with very promising results. These approaches however are complex, computationally demanding and require a large amount of data in order to train the classification models.

On the other hand, most buildings nowadays possess some form of digital documentation, with Building Information Model, or BIM, being the most prominent one. This documentation contains vast information about the building's geometry and diverse characteristics, which can be leveraged in order to obtain semantic information for the navigation of a mobile robot. A previous work from our group (\cite{karimi2021ontologybased}) proposed an ontology that allows the extraction of BIM information to be used by the Robot Operating System (ROS). This solution, named Building Information Robotic System (BIRS), leverages BIM semantics translated for robot indoor navigation. This semantic information can be provided to the robot as an a priori map to support localization and, intertwined with the robot navigation and mission, can help an operator manage the robotic system from a shared conceptual knowledge of the environment (\cite{kostavelis2017semantic}).

This paper proposes a novel method for semantic robot navigation based on an optimal path planning algorithm in two levels : global and local planning. This cascade design uses building knowledge, in the form of both metric and topological maps, extracted from BIM. The information is selected based on the previous work by \cite{karimi2021integration}. The cascading path planner generates an optimal path to navigate to destinations chosen by a user operating the system remotely. The resulting path (which is not necessarily the shortest path) can be altered according to several criteria such as robot and people safety, quality of the scan data in each location and sensors sensitivity to known environmental features. Furthermore, all along the mission, the local paths are computed using relevant complementary information for the low-level navigation (obstacles, dangers, etc). For instance, a robot should avoid getting near glass walls: they are harder to detect by many sensors. 

On the other hand, as the robot navigates, new obstacles that are not be in the BIM-extracted map can be detected, such as moved furniture. The localization and navigation can be improved if the robot is able to register these new elements and update the map accordingly. To serve that purpose, our solution integrates an active equipment detection strategy and a probabilistic map merging algorithm fusing the trustful information (onboard sensing or BIM).

The current paper contributions are as follows:

\begin{itemize}
    \item An optimal high-level path planner integrated with the low-level navigation (cascade navigation stack);
    \item A flexible semantic teleoperation and navigation for an autonomous UGV in human made structures;
    \item A practical implementation of the whole system, including active equipment detection, deployed on an autonomous mobile robot navigating a construction site.
\end{itemize}

The work is organized as follow. Section \ref{sec:related_work} summarizes the inspirational works to our approach. Section \ref{sec:topological} describes the generation of topological maps (hypergraphs) from BIM information. Section~\ref{sec:path} details the path planning algorithm. Section \ref{sec:map_merging} explains how onboard maps are merged to the semantic map. Section \ref{sec:uwb} describes our equipment detection approach based on Ultra-Wide Band (UWB) localization. The simulations and field deployment used to validate the proposed system are described in section~\ref{sec:experiments}. The results of our experimental validation are discussed in section~\ref{sec:results}. Finally, section~\ref{sec:conclusion} summarize the contributions and the next steps of our work.

\section{Related work}
\label{sec:related_work}

Conventional methods of indoor path planning often refer to the optimal path as the shortest path calculated by various algorithms such as A* and Dijkstra's (\cite{palacz2019indoor}). However, short and safe (avoiding collisions) attributes are often not enough to address complex navigation tasks. In human-made infrastructure, Building Information Modeling can provide the robot with elements' semantics and geometries (\cite{karan2015extending}) increasing its understanding of the mission. Many studies suggested ways to leverage BIM for indoor navigation. \cite{wang2020bim} develop a framework to convert the BIM digital environment to a cell-based infrastructure to support indoor path planning. In their work, they emphasize on the \textit{"BIM voxelization"} process leaving the path planning problem unaddressed. In \cite{song2019construction}, a BIM-based path planning strategy is used for equipment transport on construction sites. The authors extract the start and end points from BIM and then generate the shortest sequence of rooms for the operator, but do not support low-level adaptive robot path planning. \cite{ibrahim2017interactive} propose a path planning strategy based on BIM for an Unmanned Aerial Vehicle (UAV) on construction sites but only for outdoor usage. \cite{follini2020combining} proposed a UGV-based logistic support system with a human-assisted approach enhanced by a metric map extracted from BIM, but all experiments were performed in a controlled environment and the BIM semantics are not thoroughly leveraged. In \cite{ibrahim20194d}, the optimal route for a data collection mission using an UAV is proposed, leveraging 4D BIM to identify which building spaces are expected to change, but implemented only in a simulated environment.

\cite{delbrugger2017navigation} developed a framework supporting humans and autonomous robots navigation which mostly uses building geometries in a simulated environment. In \cite{nahangi2018automated}, the indoor localization of an UAV is assessed using calibrated visual markers (known location in the BIM).The markers setup is a tedious step that will not fit most deployment scenarios. Another study focused on the use of BIM for robot localization with hierarchical reasoning for path planning \cite{siemikatkowska2013bim}. Along that line, BIM was also demonstrated to be powerful for the identification of various paths through a graph strategy \cite{hamieh2020bim}. An approach using hypergraphs generated from IFC files was also developed in which a modified A* algorithm is able to detect the optimal path among the rooms \cite{palacz2019indoor}. These works provide only high-level path (rooms sequence) with respect to BIM geometries and lack the integration with a standard robotic architecture.

For active equipment detection, we extend the vast literature on UWB usage for localization. For instance, \cite{yin2016uwb} presented an indoor positioning system, based on the UWB technology. That work applied the trilateration localization algorithm to find the position of a \textit{blind node} based on range measurements and known UWB anchors' positions. The principles of the algorithm are transferable to our use case where we are doing the exact opposite, i.e, finding the position of the anchors based on range measurements and the position of the \textit{blind node}. The beacon localization case has also been studied by \cite{sato2019rapid} where they localized BLE beacons using RSS observations. For the localization, a person moves around the room to obtain several RSS observation. Like us, they use a range-only and inter-beacon relative distance of twin beacons approach but without a priori information, so the employ a full EKF-SLAM algorithm in order to obtain the person's location.

In these inspiring works three aspects of the domain knowledge potential for indoor robot path planning are yet to be thoroughly studied: (1) considering the full potential of the building semantic rather than only the geometry (2) integrating the high-level (rooms sequence) with the low-level sensor-based information in a synergistic navigation stack (3) validating in the field the performance of using semantic information. In this paper, we cover these gaps by integrating Building Information Robotic System (BIRS) into a navigation system in ROS for autonomous navigation and intuitive teleoperation.

\section{BIM-based topological maps}
\label{sec:topological}

Topological maps are a conventional method for robot navigation that can be fed with building-related data. One can create a graph of rooms, doors and corridors with BIM-based geometric and semantic information \cite{strug2017reasoning}. Since BIM also includes a lot of information useless to the robot, the relevant data must be identified, extracted and translated \cite{karimi2021ontologybased}. In this paper, the hypergraph of Palacz et al. \cite{palacz2019indoor} is extended to incorporate Industry Foundation Classes (IFC) semantics and geometry. The information is captured in the form of a topological map to be translated to the robot.

Nodes and edges of the graph are populated with the information provided by Building Information Robotic System (BIRS) \cite{karimi2021ontologybased}. Nodes contain room's name, room's unique ID, room center, room area, walls' unique IDs, wall material, last scan date, construction activity (hazard for the robot) and edges contain door's unique ID, door's location, doors opening direction.

\begin{figure}[!t]
    \centering
	\includegraphics[width=\linewidth]{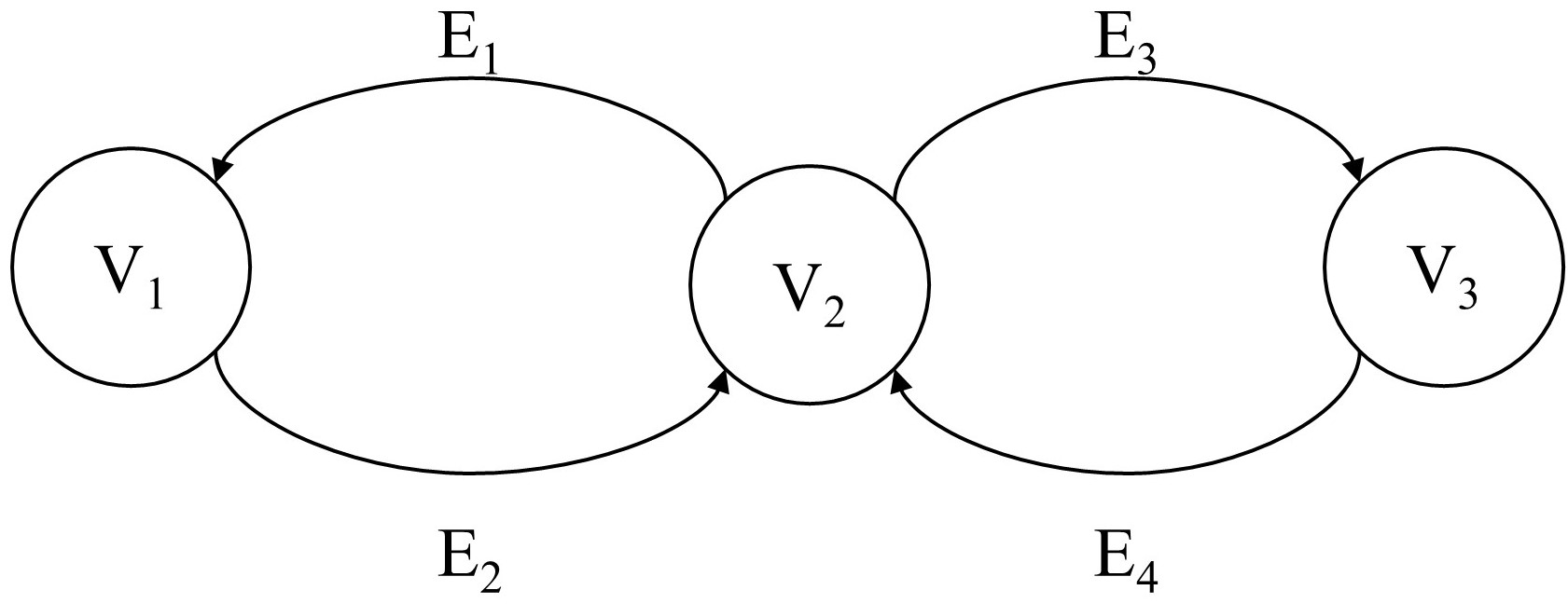}
	\caption{A directed hypergraph of $S = (V, E)$ where $V=\{V_1,V_2,...,V_n\}$ is a set of nodes and $E=\{E_1,E_2,..,E_m\}$ is a set of hyperedges. Each node $(V_i)$ is an \textit{IfcSpace} containing its relationships and each hyperedge $(E_j)$ is an \textit{IfcDoor} with its attributes extracted by BIRS \cite{karimi2021ontologybased}.}
	\label{fig:hypergraph}
	\vspace{-1.5em}
\end{figure}

In the hypergraph, one node is created per \textit{IfcSpace} containing \textit{IfcWall} and \textit{IfcCurtainWall} classes. Then, the cost of navigating between nodes is assigned to corresponding hyperedges. That is, $W=(W_V, W_E)$ is a pair of weights where $W_V$ and $W_E$ are the node and hyperedge weights respectively. $W_{V_i}$ is the $i$ node total weight obtained from:
\begin{equation}\label{W_V}
    W_{V_i}=w_{m_i}+w_{a_i}+w_{s_i}+w_{h_i}
\end{equation}
where $w_{m_i}$ depends on the walls material, $w_{a_i}$, on the room area, $w_{s_i}$, on the room scan-age, and $w_{s_i}$, of the hazards in room $i$. $W_{E_j}$ is the $j$ hyperedge weight depending only on the door opening direction: it is assumed that pulling is harder than pushing for a UGV. For passing from one node to the other, there might be several paths available. The overall weight of a path (from start to end node) is computed as follows:
\begin{equation}\label{W}
    W = \sum_{i=1}^{n}W_{V_i} + \sum_{j=1}^{m}W_{E_j}
\end{equation}
A path with lower weight is better. To help the robot avoid potential sensor erratic behavior, the material properties of the walls are included in the hypergraph through \textit{IfcMaterial} and its super-type \textit{IfcProduct}. The weight of each curtain wall, i.e. \emph{invisible} walls, in each node is $w_m=12$, and all others are $w_m=4$ since they can be easily detected. To include the passing time from one node to the other, rooms area are also included in the hypergraph, i.e. bigger rooms take more time for the robot to cross. Accordingly, the weight for the rooms less than $50m^2$, between $50m^2$ to $100m^2$ and more than $100m^2$ are $w_a=2$, $w_a=8$ and $w_a=12$ respectively. To increase the efficiency of data collection, the nodes are attributed a scanning age  (time since the last scan) to maintain the building knowledge up-to-date. It may be beneficial for the robot to visit more rooms and collect more data. Industry needs define the frequency of data collection, therefore, we assign $w_s=10$, $w_s=6$, $w_s=0$ for the scanning period of less than 1 week, between 1 week and 2 weeks, and more than 2 weeks respectively. To consider robot's safety while navigating, each node is also attributed with risk level (i.e. construction activity). For instance a high risk would be attributed $w_h=500$. In this case, any other available path should be selected by the algorithm. If there is not an alternative safe path for the robot, the algorithms provides a warning for high-weight paths so that the supervisor of the robotic deployment is warned. We use directed hypergraph (with directed hyperedges) allowing us to assign cost for door opening directions. \textit{IfcDoor} as a sub-class of \textit{IfcBuildingElement} provides the center coordinates of the doors creating hyperedges (with their coordinates) in the hypergraph. \textit{IfcDoor} also stores the opening direction through y-axis of \textit{ObjectPlacement} parameter. For pushing and pulling the door, we assign $w_d=2$ and $w_d=6$ to the hyperedge's weight respectively.

\section{Finding the optimal indoor path}
\label{sec:path}

\begin{figure}[!t]
    \centering
\noindent\begin{boxedminipage}{\linewidth}
    \begin{alltt}
Inputs:
  layout\_graph : hypergraph
  tail\_room, head\_room : node
  door : hyperedge
  path_weight : hyperedge\_total\_weight\\
Outputs:
  semantic_path : list<node, hyperedge>
  x\_y\_path : list<nodes\_coordinates,
  hyperedges\_coordinates>
  hyperedge\_total\_weight : number
    \end{alltt}
\end{boxedminipage}
    \caption{Data structure for IFC-based semantic optimal path planner algorithm}
    \label{fig:data_structure}
	\vspace{-1.5em}
\end{figure}

\begin{figure*}[!t]
	\centering
	\includegraphics[trim={0 1.75cm 0 1cm}, clip,width=\linewidth]{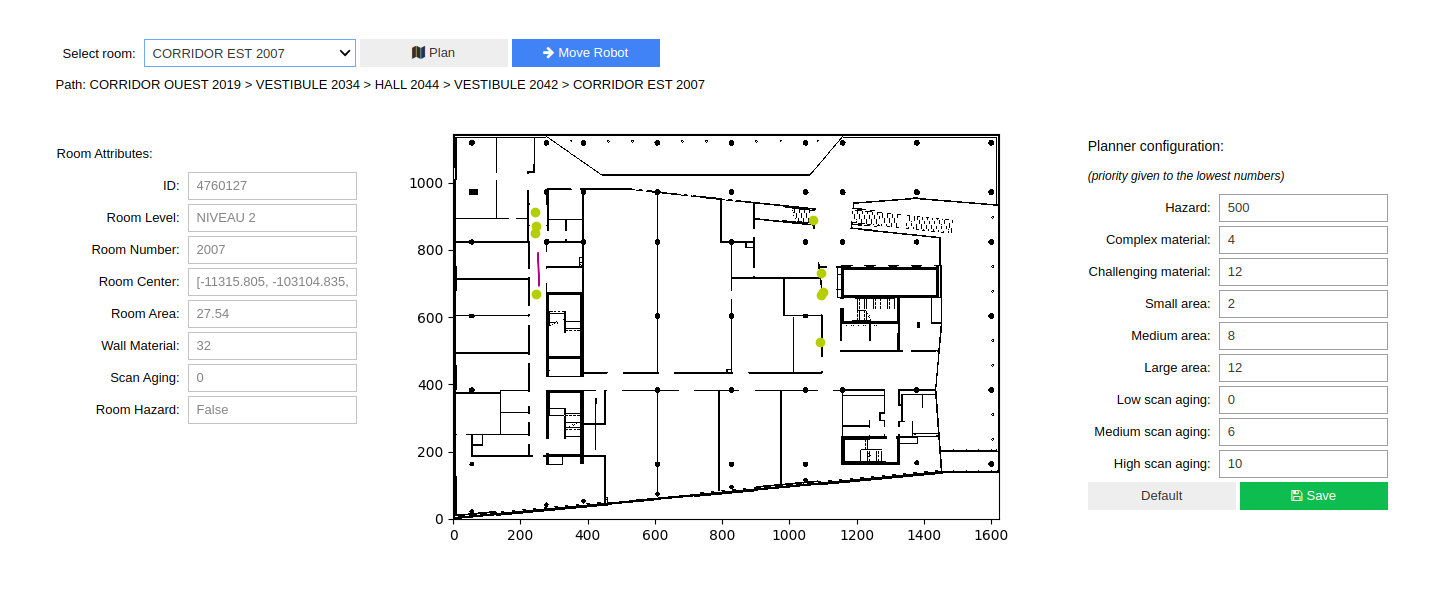}
	\caption{Semantic Graphical User Interface for the intuitive operation of a UGV with domain knowledge. The controls in the header allow selecting a destination and generating the path. The panel to the left shows the attributes of the selected room. The center contains a map of the environment, with the robot's pose in real time being represented by the purple arrow. The center points of the rooms and doors in the path are represented in the map by the yellow circles. The right panel allows the user to reconfigure the different weights that are applied to the path generation.}
	\label{fig:gui}
	\vspace{-1.0em}
\end{figure*}
The process of generating the hypergraph is done with a Dynamo Script (a visual programming tool) to extract the IFC classes and their parameters. Then, the information is stored in a XML database to be retrieved by a Python script. In order for the data to be ROS-friendly, the Python script translates the data. Having the hypergraph of the building, start and end nodes (rooms) can be defined by the user and the optimal path is automatically generated by the algorithm. We implemented directed BF-hypergraph since each room (node) can be connected to multiple rooms \cite{gallo1993directed}. Since door opening direction is considered for assigning weights to the hyperedges, each pair of nodes is connected forward and backward. This creates backward and forward sub-hypergraphs within the overall topological map. \textit{"Shortest Sum B-Tree"} is used to find all the possible hyperedges between the start and end nodes. The hyperedges contain the weight of passing from one node to the other. Having a set of paths from start to end nodes with their weights enables the \textit{"Shortest Sum B-Tree"} to select the path with the lowest cumulative weight. Since the hypergraph represents the building semantics, all the information for semantic navigation is provided to the user. As illustrated in fig. \ref{fig:data_structure}, the optimal path outputs a set room names, their coordinates and a set of door coordinates in the sequence of node location and hyperedge (door) location.

\section{Probabilistic map merging with semantic data}
\label{sec:map_merging}

The previous step can feed a standard onboard navigation stack to plan a path using a LiDAR, the known layout, and an algorithm such as A*. However, that would leave out all the semantic knowledge still available from the building model. In order to properly locate the robot in the building layout while updating that layout from the mission measurements, we developed a probabilistic map merging algorithm that takes into account the semantic data. It is composed of a localization module based on a particle filter that finds the robot's pose in the a priori map using a laser scan probabilistic model that changes the expected sensor values based on the semantic classes of the objects in the environment. In this section we explain the implementation in detail.

We consider a robot moving in the 2D plane with pose $x_{t}$ at time $t$ given by:

\begin{equation}
    x_{t} = \begin{pmatrix}x & y & \theta\end{pmatrix}^{T}
\end{equation}

The environment is represented by a map $m$ as a grid of cells $m_{i}$, $i \in \{1,...N\}$ where $N$ is the number of cells. Each cell has an associated tuple $(p_{i}, c_{i})$ where< $p_{i} \in [0, 1]$ is the occupancy value of that cell and s$c_{i} \in \{0, 1,...,C\}$ is an integer representing the class to which that cell belongs. For occupied cells, the class refers to the material that surface is made of, as identified by a robot equipped with the necessary sensors. If the material can't be identified, we use the value 0, same for unoccupied cells.

Assuming we start with an a priori map extracted from BIM, we first want to estimate the pose of the robot relative to the map's reference frame. Similar to the Monte Carlo Localization algorithm described in \cite{thrun2002probabilistic}, we construct a particle filter where each particle $(i)$ is a tuple $(x_{t}^{(i)}, w_{t}^{(i)})$ such that $x_{t}^{(i)}$ is the robot pose and $w_{t}^{(i)}$ is the particle weight at time t. For each control/measurement pair, the particles are updated according to the following equations:

\begin{equation}
    \begin{split}
    &x_{t}^{(i)} = M(u_{t-1},x_{t-1}^{(i)}) \\
    &w_{t}^{(i)} = S(z_{t},x_{t}^{(i)},m)
    \end{split}
\end{equation}

Where $M$ is a motion model and $S$ is a measurement model, as defined in \cite{thrun2002probabilistic}. To mitigate misdirection caused by inconsistencies between the map and the sensor data we modified the sensor model to account for the class information associated with each cell in the map. This information is encoded in the map when the semantic data is extracted from BIM. Below we describe our measurement model:

The measurement model is defined as a conditional probability distribution $p(z_{t}|x_{t},m)$, where $x_{t}$ is the robot pose, $z_{t}$ is the measurement at time $t$ and $m$ is the map of the environment.

For a 2-D laser range finder with $K$ beams, $z_{t}$ is a collection of measurements $z_{t}^{k}$ where $k \in \{1,..., K\}$.

The probability $p(z_{t}|x_{t},m)$ is obtained as the product of the individual measurement likelihoods:

\begin{equation}
    p(z_{t}|x_{t},m) = \prod_{k = 1}^{K}p(z_{t}^{k}|x_{t},m)
\end{equation}

\begin{figure}[!t]
	\centering
	\includegraphics[width=\linewidth]{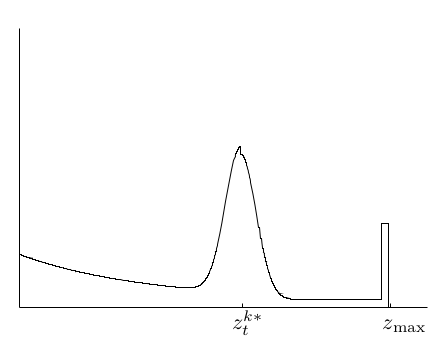}
	\caption{Probability density for a beam of a 2D laser scan sensor. $z_{max}$ is the maximum sensor range and $z_{t}^{k*}$ is the theoretical true value obtained from a ray casting algorithm, as shown in \cite{thrun2002probabilistic}}
	\label{fig:sensor_model}
\end{figure}

A typical laser range finder model is a mixture of four densities, each corresponding to a different error the sensor measurements are subject to. Figure \ref{fig:sensor_model} shows a visualization of the resulting density, where $z_{max}$ is the maximum sensor range and $z_{t}^{k*}$ is the true value that would be measured by beam $k$ in a ideal world, given the robot pose $x_{t}$ and the map $m$. This value can be obtained by a ray tracing algorithm, such as Bresenham’s line algorithm \cite{bresenham1965algorithm}, which draws a line starting on the robot's pose and moving outward, according to the beam's angle. Once an occupied cell is found, it returns the distance between the robot and that cell. In our model, we also take into consideration the class information: the algorithm only returns if the occupied cell also belongs to a class that can be detected by the sensor. This allows the estimator to ignore obstacles that wouldn't be detected by the current sensor, like a glass wall and a laser scanner. 

Once the robot's pose is obtained, the current sensor readings can be combined to the occupancy map by applying the Bayes filter described in \cite{thrun2002probabilistic}. To simplify the calculation and avoid instabilities, we use the log odds notation. 

\begin{equation}
    L(m_{i}|z_{1:t}) = L(m_{i}|z_{1:t-1}) + L(m_{i}|z_{t})
\end{equation}

The actual occupancy probabilities can be easily recovered using:

\begin{equation}
    L(m_{i}|z_{1:t}) = log \dfrac{p(m_{i}|z_{1:t})}{1 - p(m_{i}|z_{1:t})}
\end{equation}

\section{UWB-based equipment localization}
\label{sec:uwb}

While the robot moves around and update its map, it can detect active equipment that need to be tracked on site. Our solution to this challenge requires ranging measurements performed between the robot and the beacons placed on the equipment. We selected UWB technology for our beacons: a radio technology that uses a very low energy level for short-range communications and that operates at high frequencies (3.1 to 10.6 GHz). That choice was mainly motivated by the accuracy of the UWB technology and its robustness to occlusions. In fact, according to \cite{jimenez2017finding}, it has a 10 cm precision while the BLE can go up to 5 meters.

We are thus trying to locate UWB beacons attached to equipment based on known robot positions. To do so, we adapted the concept of trilateration localization algorithm coupled with an optimization algorithm.

More commonly, trilateration serves to localize a robot based on known anchors position as presented in \cite{yin2016uwb}. We exploit the same algorithm in doing the exact opposite for our use case, i.e., localize anchors based on known robots positions.

In our use case, the robot is assumed to be moving on a 2D world, hence, the localization is done in 2.5D (with a known height for the anchors). To have a noise resistant system we used 70 known robot positions spread around the anchor. We then get a system of equation that can be solved in matrix form with a pseudo-inverse minimizing the root mean square error:

\begin{equation}
    \textbf{x} = (A^TA)^{-1}A^Tb
\end{equation}

With

\begin{equation}
    A = \begin{bmatrix}
    2(x_n - x_1) & 2(y_n - y_1)\\
    2(x_n - x_2) & 2(y_n - y_2)\\
    ... & ...\\
    2(x_n - x_{n-1}) & 2(y_n - y_{n-1})
    \end{bmatrix}
\end{equation}

and 

\begin{equation}
    b = {\tiny \begin{bmatrix}
    r_1^2 - r_n^2 - x_1^2 - y_1^2 - z_1^2 + x_n^2 + y_n^2 + z_n^2 - 2z_{a}(z_n-z_1)\\
    r_2^2 - r_n^2 - x_2^2 - y_2^2 - z_2^2 + x_n^2 + y_n^2 + z_n^2 - 2z_{a}(z_n-z_2) \\
    ... \\
    r_{n-1}^2 - r_n^2 - x_{n-1}^2 - y_{n-1}^2 - z_{n-1}^2 + x_n^2 + y_n^2 + z_n^2 - 2z_{a}(z_n-z_{n-1})
    \end{bmatrix}}
\end{equation}

With the first approximation obtained from the trilateration, we use the \textit{Trust Region Reflective} algorithm to refine the solution. The algorithm solves a minimization problem based on a given cost function and an initial guess for the solution. The cost function here is the sum of the differences between the range measured by the UWB device and the same distance based on the estimated solution, i.e.:
\begin{equation}
error = \sum_{i=1}^n |r_i - (\lVert \textbf{x} - \textbf{robot}_i\rVert) |
\end{equation}
with\\
$r_i$: the range measured by the UWB device at t = i,\\
\textbf{x}: the solution tested by the minimization algorithm,\\
$\textbf{robot}_i$: the robot position at t = i.

\section{Semantic graphical user interface}
\label{sec:gui}

A Graphical User Interface (GUI) was developed based on BIM semantics to allow users to intuitively operate the robot and configure the path planner. The GUI connects to the ROS running in the robot and presents semantic information of the building and data from the robot in real time. The integrated high-level and low-level navigation system moves the robot to the desired destination. The GUI allows the non-expert users to work with their domain knowledge, thereby making robot deployment more intuitive and simpler. Figure \ref{fig:gui} illustrates the interface window. The GUI is developed in Python notebooks, allowing for easy integration of visualization widgets and customization.

The GUI provides the building's rooms in a drop-down list, from which the user selects a destination and then launch the path planner to find the optimal path. The center area of the GUI shows a map of the building, with the robot's pose being updated in real time, along with the paths objectives. The left panel shows the selected room's (end node) attributes. The right panel allows the user to alter the weights of each parameters of the path planner. After changing and saving the new weights, the user can generate the path again and see the results on the map. Finally, the user can click on the \emph{Move Robot} button to trigger the robot to start moving.

\section{Experiments}
\label{sec:experiments}

We validated our approach in simulation and with an experimental case study, where the system was implemented on a real mobile robot deployed to the field. The experiments were designed to allow the robot to navigate through the corridors of a building while recording data from its sensors. We choose one of the buildings at École de Technologie Supérieure, for which a complete BIM model was available. An operator commands the robot by choosing destinations in a semantic graphical user interface running on a computer, which then sends this destination to the robot through a wifi link. The hierarchical path planner running on the robot generates a list of waypoints to the destination and an internal controller moves the robot accordingly. In the next sessions we describe each of those parts in detail.

\subsection{Robotic platform}
\label{sec:robot}
\begin{figure}[!t]
	\centering
	\includegraphics[width=\linewidth]{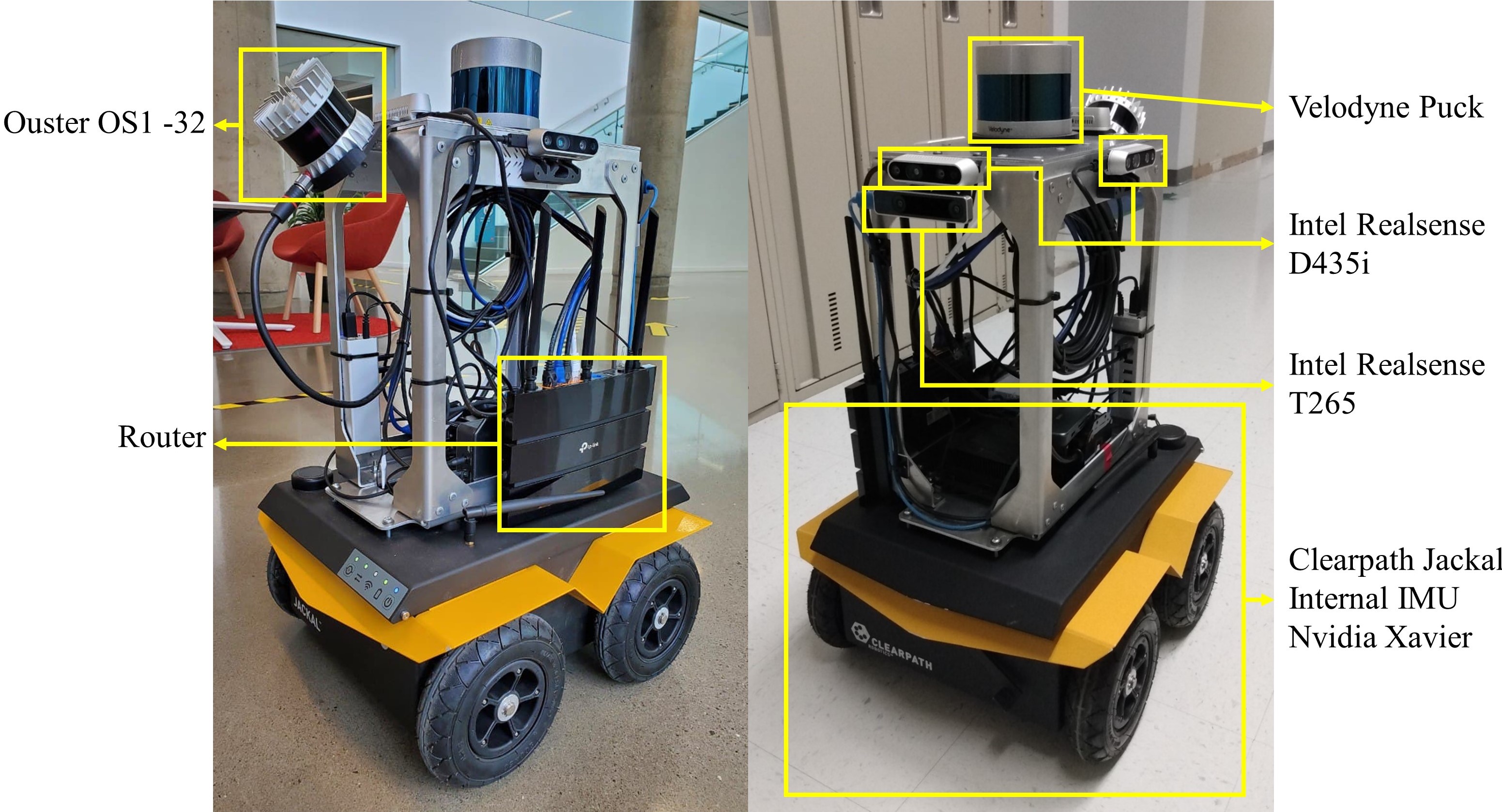}
	\caption{Mobile robot platform equipped with various sensors}
	\label{fig:robot_platform}
	\vspace{-1.5em}
\end{figure}

The robotic platform used in this study is built from a four-wheeled unmanned ground vehicle (Clearpath Jackal) equipped with a hybrid vision/laser scan sensing system and is shown in fig. \ref{fig:robot_platform}. This robotic system is capable of collecting data while the robot is moving and can navigate semi-autonomously, with a remote operator sending high level commands.

The Jackal is equipped internally with wheel encoders and an IMU, which are used for odometry estimation and control by an embedded computer. Additionally, a NVidia Jetson AGX Xavier computer was installed to process the data and the navigation algorithms. A suite of ROS nodes for control, state estimation and diagnostics are provided by Clearpath, which serve as a basis for the robot operation.

The sensing system was envisioned for data collection in construction sites and contains LiDARs and cameras positioned in different directions to cover as much of the robot's surroundings as possible. A front-facing Intel Realsense T265 tracking camera combines the mobile platform's wheel odometry with visual data and outputs high accuracy odometry estimation to the localization system. Another front-facing camera, a Intel Realsense D435i depth camera feeds depth images to a collision avoidance node that reacts when the robot moves too close towards an obstacle. A Velodyne Puck 32MR Lidar is mounted horizontally to scann the walls and floor and is used to detect obstacles and objects by the localization and map merging algorithms. The other sensors are integrated for data collection and used in other studies.

\begin{figure}[!t]
	\centering
	\includegraphics[width=\linewidth]{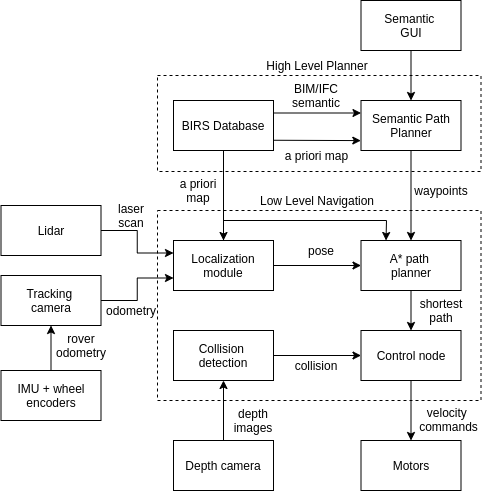}
	\caption{System Overview: A high level planner that process BIM/IFC information and user inputs is integrated to a low level navigation stack in a cascade design. The low-level module takes care of the localization, local path planning and collision avoidance tasks, while the high-level planner generates paths based on BIM/IFC semantics.}
	\label{fig:system_overview}
\end{figure}

Figure \ref{fig:system_overview} gives an overview of the system. The robot pose in the map is obtained through the use of a ROS implementation of the Adaptive Monte Carlo localization algorithm from \cite{thrun2002probabilistic}. Before deploying the robot, wall geometry information is extracted from BIM to generate an occupancy grid of the building. During the robot navigation, this map, the odometry, and the laser scan data from the horizontally mounted Velodyne LiDAR are fed to the localization algorithm, which then estimates the robot's current pose in that map. When a destination room is selected, the semantic path planner outputs the preferred path to that room as a list of waypoints, containing the center points of each room, door and corridor in the path. An A* path planner\cite{hart1968formal} then calculates the shortest path from the robot's current position to the next waypoint in the list. Velocity commands are generated from the A* path and sent to the robot's internal controller to drive it though that path.

\subsection{Simulation}
\label{sec:simulation}

\begin{figure}[!t]
	\centering
	\includegraphics[trim={1cm 5cm 1cm 5cm},clip,width=\linewidth]{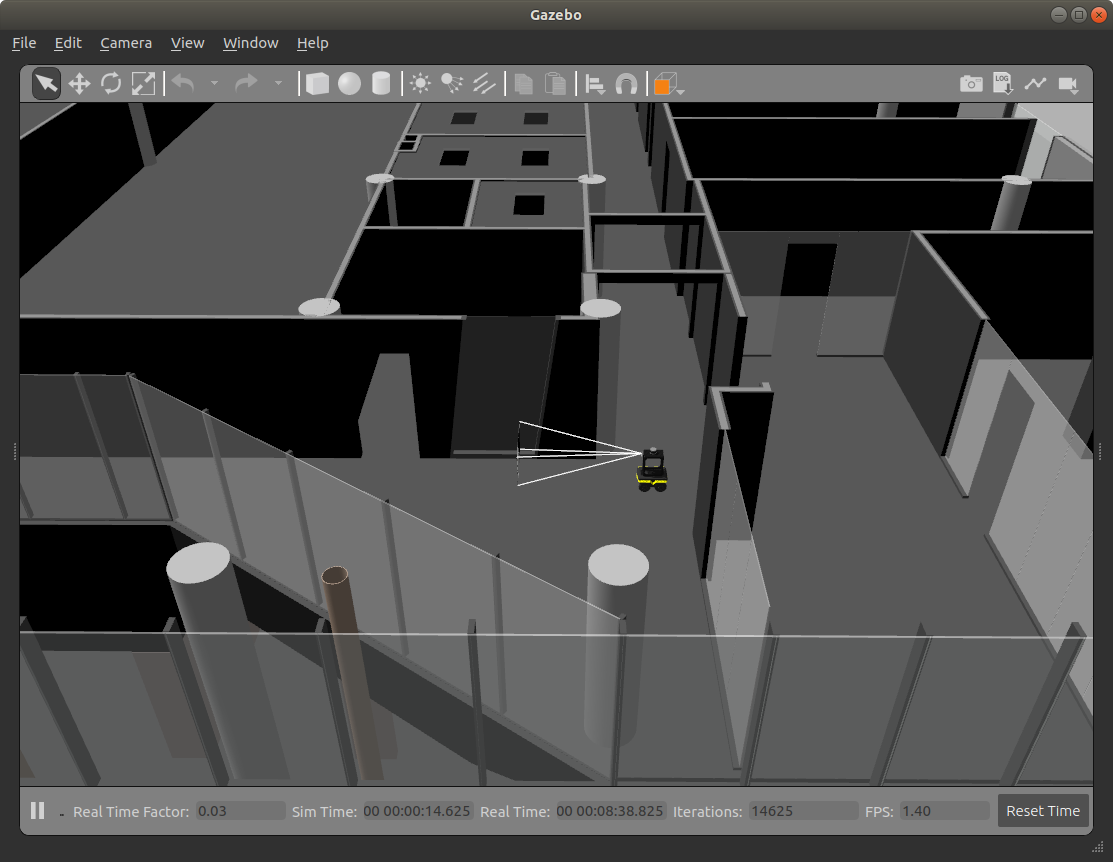}
	\caption{View of the simulated environment used to test the BIM/IFC optimal path planning approach. The building 3D model was built with geometry information extracted from the BIM. The robot model simulates the sensors and possesses the same characteristics as the real robot.}
	\label{fig:simulation}
	\vspace{-1.5em}
\end{figure}

The simulation was performed using the Gazebo Simulator. The building information is exported to create a 3D model, a digital twin. Clearpath, Gazebo and the ROS community provide all the required software packages required to generate an accurate simulation of our robotic platform. Figure \ref{fig:simulation} shows the simulated robot and its environment with different wall textures and transparency.

\section{Results}
\label{sec:results}

\subsection{UWB-Beacons Localization Validation}
\label{sec:uwb_validation}

\begin{figure}[!t]
     \centering
     \begin{subfigure}[b]{0.45\columnwidth}
         \centering
         \includegraphics[width=\columnwidth]{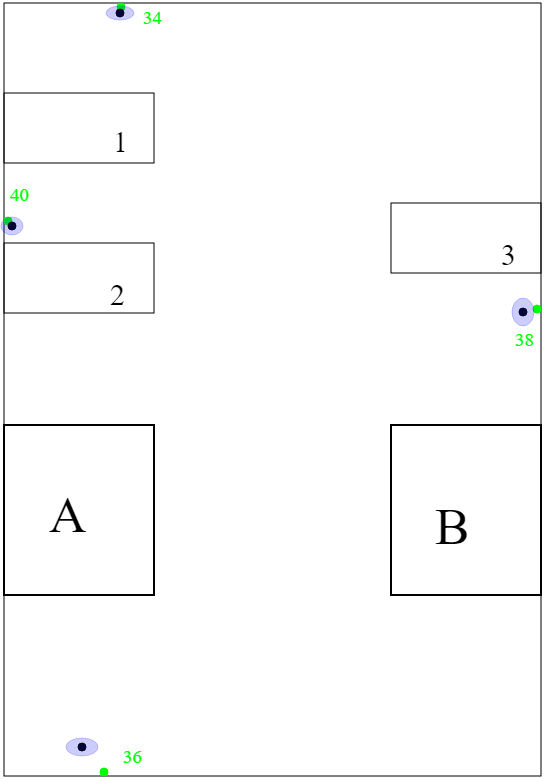}
         \caption{}
         \label{fig:mean_pos_uwb}
     \end{subfigure}
     \hfill
     \begin{subfigure}[b]{0.45\columnwidth}
         \centering
         \includegraphics[width=\columnwidth]{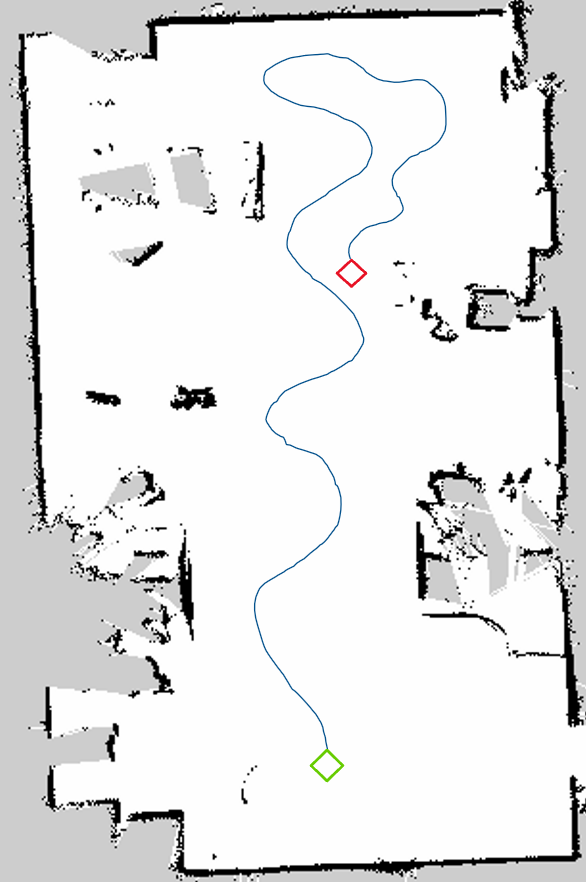}
         \caption{}
         \label{fig:trajectory_uwb}
     \end{subfigure}
     \caption{(a) Simplified room plan. The anchors are indicated with the green dots. The mean detected positions are the black dots and the blue ellipses represents the standard deviation of the measurements. A and B are 1.68m high desks and 1 to 3 are 0.70 cm high desks. (b) Robot's winding trajectory: the green lozenge represents the initial position and the red one represents the final position.}
\end{figure}

To evaluate the active equipment localization algorithm, we performed a series of tests with the Jackal robot, which was equipped with a Pozyx tag placed 78 cm above the floor level. Four UWB beacons representing equipment were positioned in the laboratory area, according to table \ref{tab:uwb_position}. Figure \ref{fig:mean_pos_uwb} shows a simplified plan of the test room where the boxes A and B are 1.65 m high desks and the boxes 1 to 3 are 70 cm high.

As discussed previously, the positions considered for the 2D trilateration have to be non-colinear, otherwise the algorithm will output erroneous results. For that reason, the robot's trajectory must be winding in order to ensure it. The algorithm also enforces a 10 cm interval between consecutive positions used in the algorithm. Figure \ref{fig:trajectory_uwb} shows an example of a winding trajectory on a generated plan of the test room.

After a set of 20 experiments (20 detections per anchor) we obtained the results presented in table \ref{tab:uwb_error}. That results show that the mean error on the detection is around 11 cm on both x and y axes. The worst results comes from the tag \textit{36} which is partially occluded by a desk. The minimal error is about 1 cm while the maximal goes up to 63 cm. Figure \ref{fig:mean_pos_uwb} also presents a graphical representation of the error and standard deviation for each tags. The green dots represent the ground truth, the black dots are the mean position detected by the algorithm and the blue ellipses represent the standard deviation of the detections. 

\begin{table}[]
\begin{center}
\caption {Test setup: UWB anchors positions} \label{tab:uwb_position} 
\begin{tabular}{ |c|c|c|c| } 
\hline
Anchors & x (m) & y (m) & z (m) \\
\hline
34 & 6.62 & 1.44 & 1.66 \\
\hline
36 & -1.11 & 1.61 & 1.68 \\
\hline
38 & 3.52 & -2.72 & 1.86 \\
\hline
40 & 4.40 & 2.65 & 1.87 \\
\hline
\end{tabular}
\end{center}
\end{table}

\begin{table}[]
\begin{center}
\caption {UWB-beacons detection errors} \label{tab:uwb_error} 
\resizebox{\columnwidth}{!}{\begin{tabular}{ |c|c|c|c| } 
\hline
  & Mean error (m) & Min error (m) & max error (m) \\
\hline
X & 0.119 & 0.0116 & 0.489 \\
\hline
Y & 0.115 & 0.0119 & 0.638 \\
\hline
Distance from the origin & 0.0684 & 0.000105 & 0.294 \\
\hline
\end{tabular}}
\end{center}
\end{table}

\subsection{Navigation}
\label{sec:deployment}

The experiment had two main objectives:

\begin{enumerate}
    \item Test the effectiveness of the semantic path planner in generating the optimal path to reach the destination, given the building information obtained from BIM/IFC.
    \item Test how changes in the building information affect the final path that is generated.
\end{enumerate}

\begin{figure*}[!t]
     \centering
     \begin{subfigure}[b]{\columnwidth}
         \centering
         \includegraphics[width=\columnwidth]{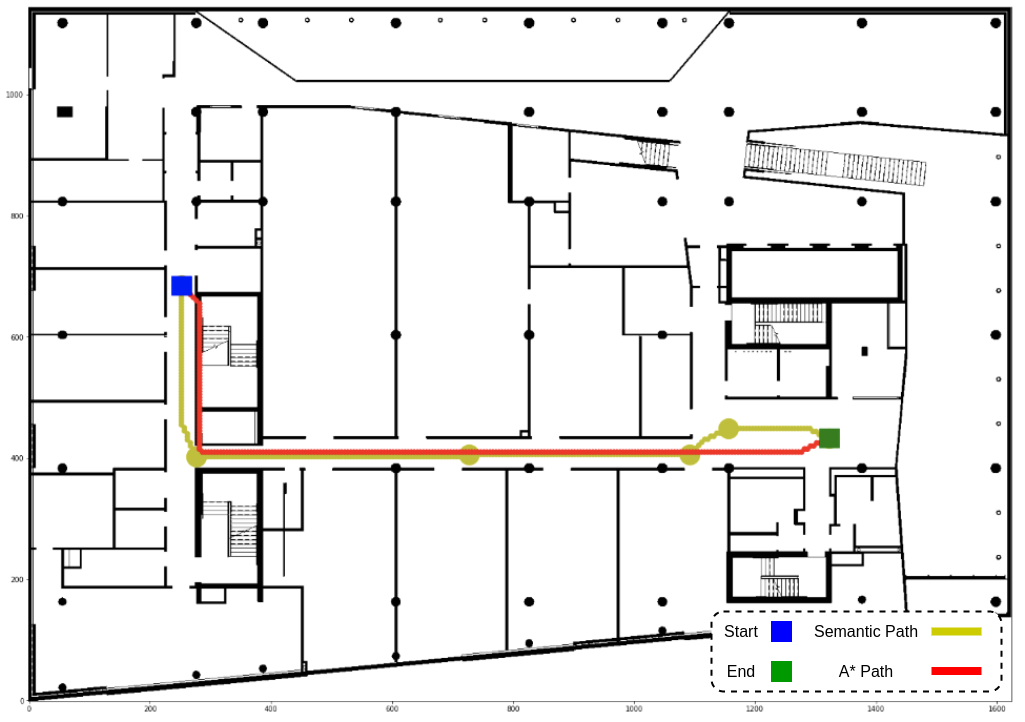}
         \caption{}
         \label{fig:astar_semantic_normal}
     \end{subfigure}
     \hfill
     \begin{subfigure}[b]{\columnwidth}
         \centering
         \includegraphics[width=\columnwidth]{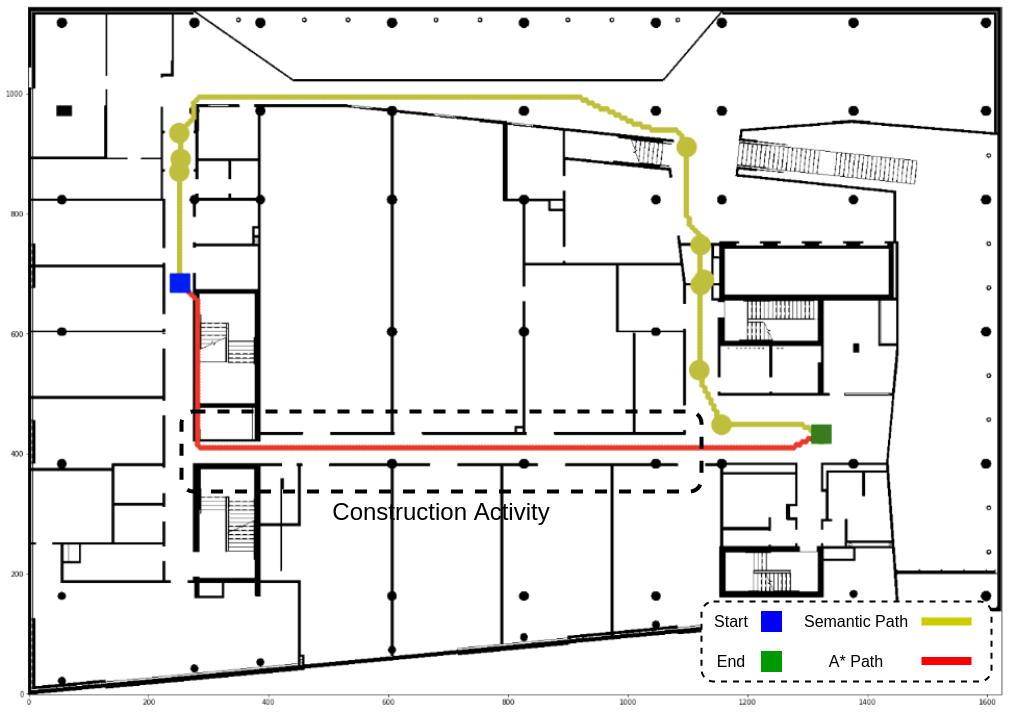}
         \caption{}
         \label{fig:astar_semantic_hazard}
     \end{subfigure}
     \caption{High-level and low-level paths: A* generates the shortest path possible between start and end, not taking advantage of the BIM/IFC semantics. (a) Without any special condition, both algorithms generate the shortest path. (b) When a construction activity is happening, our semantic path planner is able to find an alternative path, while A* causes the robot to navigate through the hazardous area.}
\end{figure*}

\begin{figure*}[!t]
     \centering
     \begin{subfigure}[b]{\columnwidth}
         \centering
         \includegraphics[width=\columnwidth]{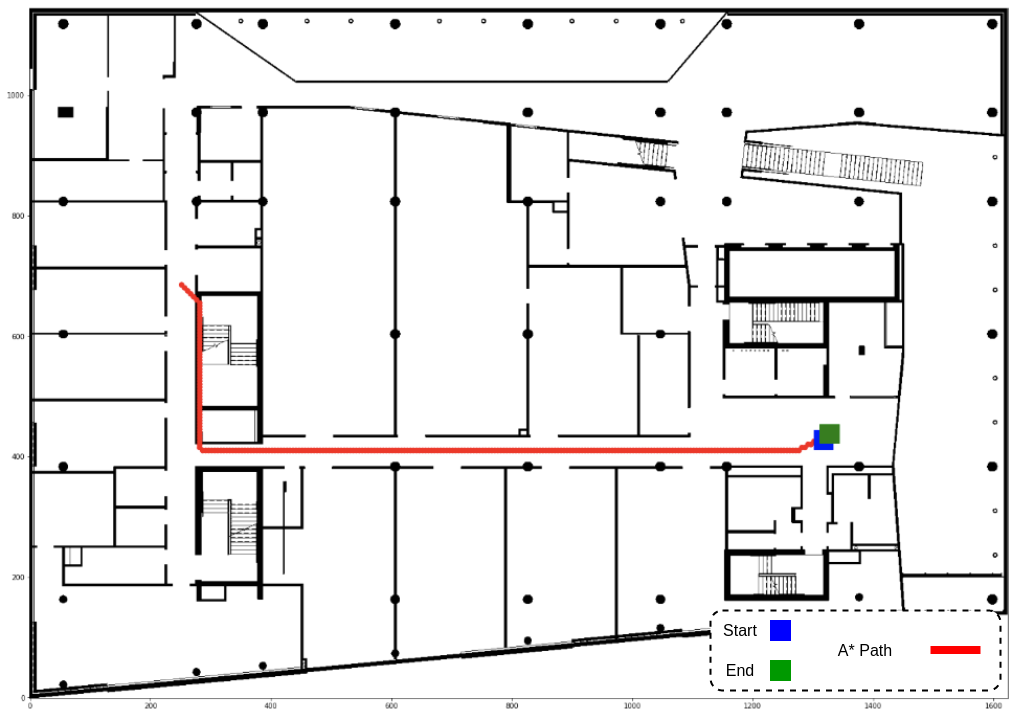}
         \caption{}
         \label{fig:astar_explored}
     \end{subfigure}
     \hfill
     \begin{subfigure}[b]{\columnwidth}
         \centering
         \includegraphics[width=\columnwidth]{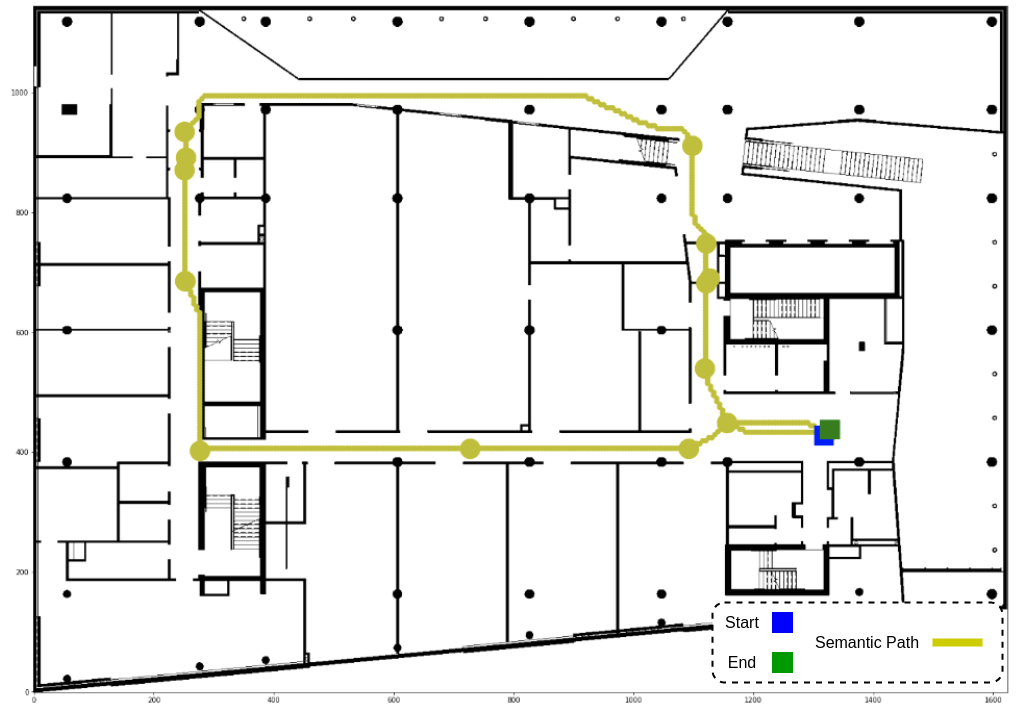}
         \caption{}
         \label{fig:semantic_explored}
     \end{subfigure}
     \caption{BIM/IFC semantics can be used to increase explored area. (a) By searching for the shortest possible path, A* causes the robot to navigate the same corridor twice. (b) The semantic path planner avoids recently explored areas, resulting in paths that cover more of the environment.}
\end{figure*}

In our case study, the robot starts in a corridor (CORRIDOR OUEST) on the west side of the building and must reach an open area (CORRIDIR EST) on the eastern part of the building. Figure \ref{fig:astar_semantic_normal} shows the building map, and the path in red line generated by applying the A* algorithm from start to end. This is the shortest possible path between the two points, taking into consideration only the building geometry and a small safety collision radius around the robot. When the Semantic Path Planner is applied to the same scenario, a similar result is obtained as expected, represented by the yellow path in fig. \ref{fig:astar_semantic_normal}. Since there are no doors, undesirable materials or hazards in the path, the algorithm outputs a list of rooms that must be visited by the robot that represent the shortest distance from start to end. The semantic path planner provided the order of rooms' names from the start to the end as it is show in the GUI in fig. \ref{fig:gui}. Therefore, the user operating the robot can intuitively track the path from the data collected. In this direction, the as-built data can be directly compared to the as-planned since the path is recorded semantically. Also, the waypoints of rooms' center coordinates and doors' center coordinates are provided by the semantic path planner. If there is a door made of materials invisible to sensors (such as glass), the complementary door coordinates helps for safer, smarter, precise data collection. Following this, the A* algorithm finds the shortest path between the waypoints. 

In a second run, the building information was altered to include a construction operation carried out in the area highlighted with a dashed box in fig. \ref{fig:astar_semantic_hazard} (not visible in the GUI). Since the construction activity represents a hazard with a high cost for the Semantic Path Planner, a different path passing through another corridor is automatically selected, as illustrated by the orange path. Nevertheless, the high cost of the shortest path triggered a warning in the system indicating a hazard to the user through the semantic GUI. Therefore, the user can understand the risks associated with navigation through an active construction area and decide whether to scan the environment or postpone it to a safer time. The yellow path was automatically generated, although it is not the shortest path, as the optimal path from the default parameters mentioned in section \ref{sec:path}. This path passes along a large curtain wall invisible to the robot's sensors. The additional semantic information provided by the BIRS is given to the robot as well as the BIM occupancy grid so it contributes to collision avoidance with the wall. The GUI provides the user with the scan aging of the rooms so the user can decide which rooms to select as the destination for data collection. This allows the users to run multiple data collection mission with the robot which increases the efficiency of robot deployment on construction sites. 

For a final run, the robot was commanded to navigate to the destination and then return to the origin. As shown in fig. \ref{fig:astar_explored}, the red path generated by A* passes twice over the same corridor, minimizing the explored area. Our semantic path planner, however, employs BIM/IFC semantics to track the amount of time since a room was last visited and tries to avoid recently visited ones. Figure \ref{fig:semantic_explored} shows that the path generated when the robot is returning is different from the first one, encouraging the robot to seek unexplored zones. The result is a bigger portion of the environment is covered and more data is obtained by the robot. As illustrated, the integrated BIM-ROS information provides a cascade navigation system on construction sites enabling autonomous and accurate data collection of the spaces scanned.

\section{Conclusion}
\label{sec:conclusion}

This paper presented a semantic path planner that uses building information from IFC data schema to generate optimal paths for safe and efficient navigation of autonomous robots on job sites during the construction phase. We used the BIRS for extracting building information from IFC represented in a hypergraph structure. Path planning algorithms can then be used to calculate optimal paths in this graph given the building information. Weights are designated to each connection in the path to represent how different conditions can affect the robot's navigation and to prioritize paths with more desired characteristics. The optimal semantic path is then integrated with low-level navigation system and A* algorithm is used to calculate the shortest path within the optimal path. The effectiveness of the path planning to generate different paths given different conditions was shown in a simulated and real life case study.

This algorithm can be extended in the future to take into consideration Mechanical, Electrical and Plumbing (MEP) semantics for data collection. Different locations can be added based on the kind of information needed at a specific time of construction through the GUI in order to provide the robot more destinations to collect data. Therefore, the high-level path planning algorithm would provide a more efficient route for data collection as well as semantic navigation. Also, this paper provided semantic navigation of mobile robot on construction sites, therefore, a user study will be conducted in order to assess the usability of the semantic navigation approach.

\begin{acknowledgements}
The authors are grateful to Mitacs for the support of this research as well as Pomerleau; the industrial partner of the ÉTS Industrial Chair on the Integration of Digital Technology in Construction.
\end{acknowledgements}

\bibliographystyle{spbasic}      
\bibliography{bibliography.bib}

\end{document}